\documentclass{article}

\usepackage[english]{babel}

\usepackage[letterpaper,top=2cm,bottom=2cm,left=3cm,right=3cm,marginparwidth=1.75cm]{geometry}

\usepackage{amsmath}
\usepackage{graphicx}
\usepackage[colorlinks=true, allcolors=blue]{hyperref}

\usepackage{booktabs}
\usepackage{caption}
\usepackage{subcaption}
\usepackage{float}
\usepackage{threeparttable}

\usepackage{multirow}

\title{Using Images from a Video Game to Improve the Detection of Truck Axles}
\author{Leandro Arab Marcomini, Andre Luiz Cunha}

\begin{document}
\maketitle

\begin{abstract}

Convolutional Neural Networks (CNNs) traditionally require large amounts of data to train models with good performance. However, data collection is an expensive process, both in time and resources. Generated synthetic images are a good alternative, with video games producing realistic 3D models. This paper aims to determine whether images extracted from a video game can be effectively used to train a CNN to detect real-life truck axles. Three different databases were created, with real-life and synthetic trucks, to provide training and testing examples for three different You Only Look Once (YOLO) architectures. Results were evaluated based on four metrics: recall, precision, F1-score, and mean Average Precision (mAP). To evaluate the statistical significance of the results, the Mann-Whitney U test was also applied to the resulting mAP of all models. Synthetic images from trucks extracted from a video game proved to be a reliable source of training data, contributing to the performance of all networks. The highest mAP score reached 99\%. Results indicate that synthetic images can be used to train neural networks, providing a reliable, low-cost data source for extracting knowledge. 

\end{abstract}

\section{Introduction}

Neural networks enable engineers to tackle intricate problems, analyze large datasets, and develop intelligent systems that can automate processes, enhance decision-making, and improve overall system performance. With their ability to handle non-linear relationships and adapt to changing conditions, neural networks contribute significantly to advancing engineering practices and fostering innovation in diverse applications. These computational models, inspired by the structure and functioning of the human brain, are capable of learning and making predictions from complex data. They are employed for tasks like pattern recognition and fault detection \cite{avci2022overview}, optimization and control systems \cite{gan2023application}, and image processing \cite{scholes2022dronesense}.

Various transportation engineering applications use neural networks, ranging from energy storage systems to passenger demand forecasting and aircraft maintenance. In \cite{elebi2009LightRP}, the authors developed short-term passenger demand forecasting models for light rail services using multi-layer perceptron (MLP) models, with each time slot handled independently to eliminate significant seasonality. Artificial Neural Networks have also been used for automatic number plate recognition systems \cite{Badr2011AUTOMATICNP} and as an auxiliary factor in planning transportation routes \cite{Chernyshev2022TheUO}. In vision-based systems, Convolutional Neural Networks (CNNs) were used to propose a flexible neural network for fast and accurate road scene perception \cite{mehtab2022flexible}, which is crucial for autonomous vehicles. CNNs were also used to detect and count truck axles in images \cite{marcomini2023truck}, an essential task in characterizing the truck fleet for planning and operating transportation systems.

CNNs, commonly used in image classification and object detection, require vast amounts of data to extract knowledge from \cite{hensman2015impact}. An image dataset is critical for the training process of CNNs because it provides the network with patterns and features to learn from. Therefore, data collection is a crucial step in the training process of convolutional neural networks \cite{zhang2022pavement}.

Data collection involves three main steps: acquiring, labeling, and improving existing data or models \cite{Coleman2022ImageSA}. Data acquisition involves collecting raw data from various sources, such as cameras, sensors, or existing databases. Data labeling involves annotating the collected data with relevant labels or tags to provide the network with ground truth information. Improving existing data or models consists in refining the collected data or models to improve their quality and accuracy. Data collection is a challenging task that requires significant effort and resources \cite{roh2019survey}. 

\begin{figure*}[h]
	\begin{minipage}[t]{\textwidth}		
		\centering
		\includegraphics[width=0.585\textwidth]{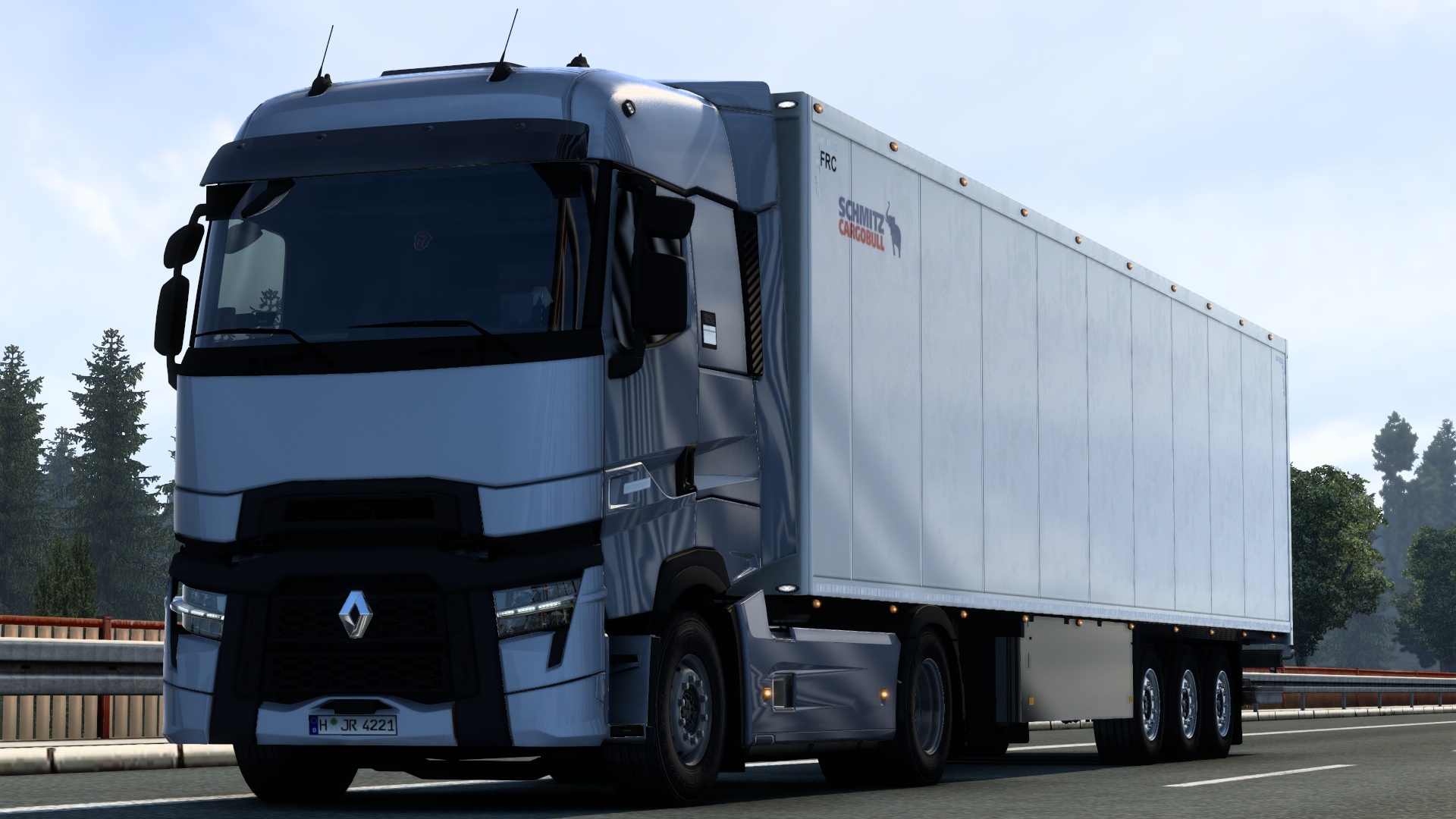}
		\label{fig:sub:subfigure2a}
	\end{minipage}
	\begin{minipage}[t]{\textwidth}
		\centering
		\includegraphics[width=0.585\textwidth]{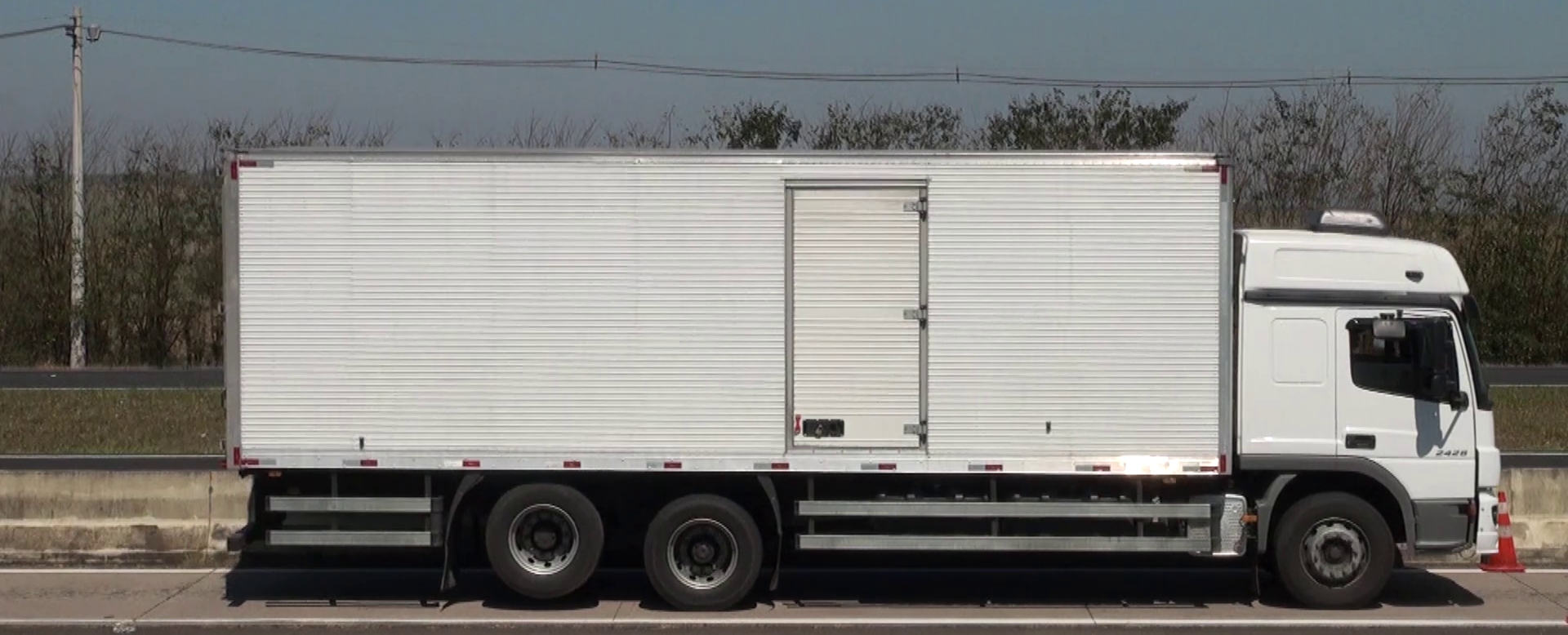}
		\label{fig:sub:subfigure2b}
	\end{minipage}
	\caption{Example of synthetic and real-world data. The truck on the top is a rendered 3D model. The truck on the bottom is a real-life truck.}
	\label{fig:synthetictrucksxreal}
\end{figure*}

To overcome the issue of limited data availability, there are different approaches for training deep networks. These include data augmentation, where the existing data is modified through various image transformations, or using pre-trained networks that have already been trained on similar tasks as a starting point for the original problem \cite{sajjad2019multi}. Another approach is to create artificial training samples using computer-generated (CG) images. This involves using computer models to generate synthetic data, which can be added to the training set, providing more diverse data for training the network \cite{yan2019data}. An example of synthetic data compared to real-world data can be seen in Figure~\ref{fig:synthetictrucksxreal}.

The main objective of this paper is to examine whether synthetic images from a video game can be effectively used to train a CNN for detecting real-life objects. To accomplish this, 27 different neural networks were trained using a combination of synthetic and real-world images of trucks, with a focus on detecting truck axles. All the networks were based on the YOLO model but were trained using different combinations of images and base weight structures. The results were evaluated using four commonly used metrics in neural networks: (1) precision, (2) recall, (3) mAP, which considers the accuracy of the object's location prediction, and (4) F1-score, a balanced metric of precision and recall.

\section{Convolutional Neural Networks}

\begin{figure}[h]
\centering
\includegraphics[width=0.6\textwidth]{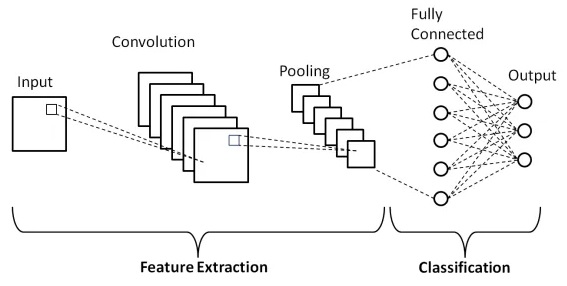}
\caption{CNN architecture, with feature extraction happening on the convolution layers, and classification provided by a fully connected MLP. Source: \cite{cnnarq}.}
\label{fig:cnn}
\end{figure}

Convolutional neural networks are a class of deep neural networks that are particularly well-suited for classifying images \cite{bhatt2021cnn}. CNNs use an ad-hoc architecture inspired by our understanding of processing within the visual cortex, and they leverage spatial information to identify patterns in images. CNNs comprise one or more layers of two-dimensional filters with activation functions and, usually, down-sampling. Each layer learns to detect different features in the input data, and the network as a whole 
can learn to abstract relevant information automatically while the data is being processed \cite{li2021survey}. In Figure~\ref{fig:cnn}, it is possible to see a common architecture of a CNN, with one convolution layer extracting features, followed by a fully connected multi-layer perceptron, known as the classification layer, to classify images based on given classes.

Several models have been developed to enhance the performance and efficiency of CNNs. One prominent model is the LeNet-5 \cite{lecun1998gradient}, which was one of the earliest CNN architectures, comprised of multiple convolutional and pooling layers. This model paved the way for more advanced architectures such as AlexNet \cite{krizhevsky2017imagenet}, which introduced the concept of deep CNNs by employing multiple layers with millions of parameters. Another notable model is VGGNet \cite{simonyan2014very}, which emphasizes the importance of using smaller filter sizes and deeper architectures. GoogLeNet \cite{szegedy2015going}, also known as Inception, introduced the concept of `inception modules' that employed multiple filter sizes in parallel, effectively capturing different scales of features. The ResNet model \cite{he2015deep} introduced residual connections to alleviate the vanishing gradient problem and enable the training of extremely 
deep networks. These models have significantly contributed to the development of CNN architectures, enhancing their capabilities in tasks such as image classification, object detection, and semantic segmentation.

An example of a modern algorithm that uses the ResNet algorithm is `You Only Look Once' (YOLO). YOLO \cite{redmon2016you} is a real-time object detection system that uses a single neural network to predict bounding boxes and class probabilities for objects in an image. YOLO divides the image into a grid, and each grid cell predicts a fixed number of bounding boxes, along with the class probabilities for each box. YOLO has gone through many development iterations, resulting in several different versions.

YOLOv3 \cite{redmon2018yolov3} uses a feature extractor based on a residual network and a spatial pyramid pooling (SPP) module to capture features at different scales. YOLOv3 also uses multi-scale training, where the network is trained on images of different sizes, to improve its ability to detect objects of different sizes.

\begin{figure}[h]
\centering
\includegraphics[width=\textwidth]{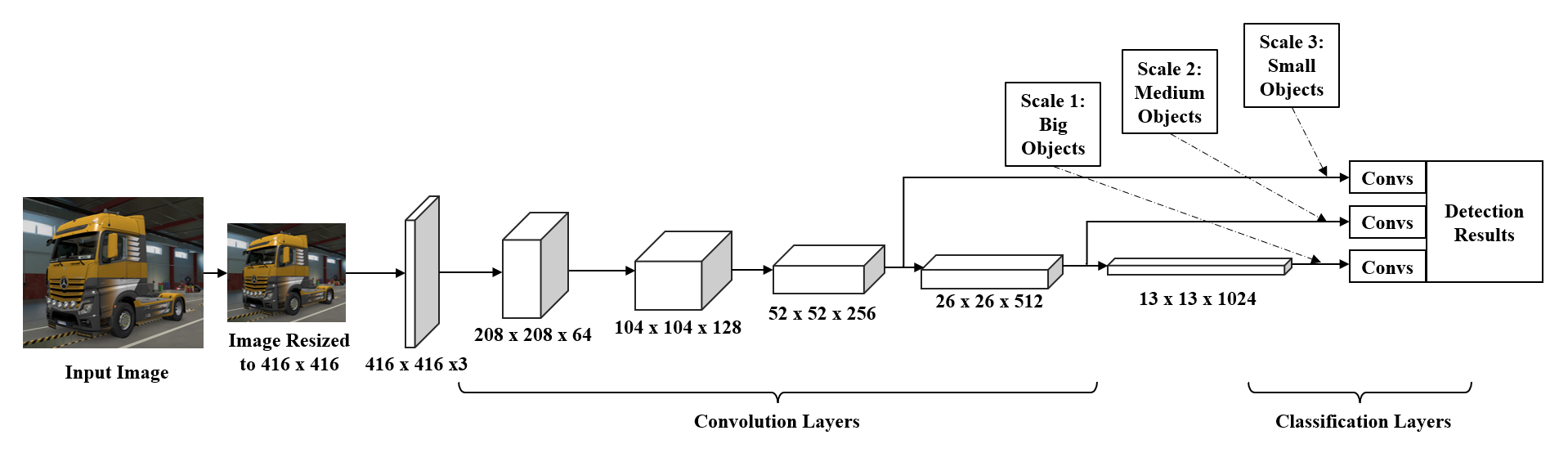}
\caption{YOLOv3 layers and architecture, with several convolution layers and different classification scales to detect different-sized objects. Source: adapted from \cite{ammar2021vehicle}.}
\label{fig:yolov3}
\end{figure}

The architecture of YOLOv3, which is the basis for all different versions, can be seen in Figure~\ref{fig:yolov3}.

YOLOv8 \cite{yolov8}, released in 2023,  uses a CSP (Cross Stage Partial) Darknet backbone with bottleneck layers, reducing the overall computational complexity from previous models. This backbone architecture has a focus layer, which splits the input channels for early feature detection, and still uses a SPP module for multi-scale detection.

YOLOv11 \cite{yolov11} was released in 2024 and introduced several architectural enhancements to improve object detection performance. The backbone incorporates C3K2 blocks, which uses smaller 3×3 convolutional kernels to optimize feature extraction and computational efficiency. It still uses an SPP module to improve the detection of objects of different sizes. Additionally, the C2PSA (Cross Stage Partial with Spatial Attention) block is integrated to improve spatial attention, enabling the model to focus on critical regions within an image \cite{khanam2024yolov11}.

Although all models follow the base structure for YOLO, the number of layers and parameters vary from model to model. A summary is presented on Table~\ref{table:yolo_summary}.

\begin{table}[h]
    \centering
    \caption{YOLO Models: Layers and Parameters}
    \label{table:yolo_summary}
    \begin{tabular}{lcc}
    \toprule
    \textbf{Model} & \textbf{Layers (Approx.)} & \textbf{Parameters (Approx.)} \\
    \midrule
    YOLOv3-Tiny  & 23               & 8.7 million          \\
    YOLOv3       & 106              & 61.5 million         \\
    YOLOv3-SPP   & 110              & 63 million           \\
    \midrule
    YOLOv8n      & 168              & 3.2 million          \\
    YOLOv8l      & 268              & 43.7 million         \\
    YOLOv8x      & 328              & 68.2 million         \\
    \midrule
    YOLOv11n     & 150              & 2.8 million          \\
    YOLOv11l     & 250              & 42.5 million         \\
    YOLOv11x     & 310              & 66.8 million         \\
    \bottomrule
    \end{tabular}
\end{table}

Since CNNs require large amounts of labeled data to learn and recognize patterns in images accurately, insufficient examples can lead to overfitting, where the network becomes too specialized to the training set of images and performs poorly on new data \cite{coleman2022image}. Because of that, data collection is crucial for neural networks because it provides the necessary information for the network to learn and make accurate predictions.

\section{Data Collection}

Data collection is gathering and measuring information on variables of interest in an established systematic fashion to systematically answer stated research questions, test hypotheses, and evaluate outcomes \cite{pauly2020expanding}. It is a critical step in research and analysis, providing the foundation for decision-making and problem-solving. Data collection can be done through various methods, such as surveys, interviews, observations, and experiments, and it can be quantitative or qualitative.

In transport engineering, data collection is essential for (1) planning and design since it helps in understanding the current state of the transport system and identifying areas that need improvement. This information is then used to plan and design new transport infrastructure and services \cite{Gulin2020BIMOT}; (2) monitoring and maintenance, where data collection is used to monitor the performance of the transport system and identify areas that require maintenance or repair. This monitoring ensures the safety and efficiency of the transport system \cite{Macchiarulo2022MultiTemporalIF}; and (3) research and development, where data collection is used to create databases to understand the behavior of the transport system and develop new technologies and solutions to improve it \cite{louro2023factors, morelli2023banco}.

\begin{figure}[h]
\centering
\includegraphics[width=.8\textwidth]{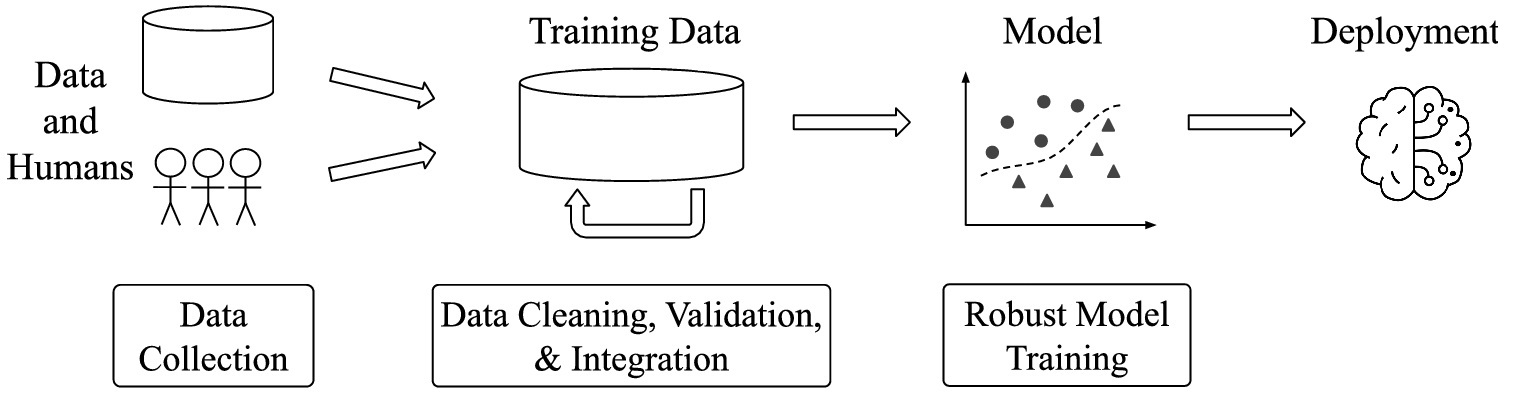}
\caption{The process of using neural networks and the importance of data collecting. Source: adapted from \cite{whang2023data}.}
\label{fig:data_collection}
\end{figure}

One way to collect data in transport engineering is by using remote sensors, such as cameras. Images can be used to analyze and detect cracks in the pavement \cite{safaei2022automatic}, to monitor urban traffic using UAVs \cite{butilua2022urban}, and to detect vehicles in traffic, to avoid collisions \cite{dong2022lightweight}. The authors commonly apply neural networks to process large amounts of data to conclude the researched subject. Figure~\ref{fig:data_collection} shows a schematic of the process. Data collection is the first step, and all subsequent steps depend on it.

Collecting real-world data can be time-consuming, expensive, and dangerous \cite{tremblay2018training}. In transport engineering, collecting vehicle images generally involves cameras mounted on the side of high-speed highways, exposed to vehicle fumes, and on different weather and lighting conditions \cite{arora2022automatic, marcomini2023truck}. That is why synthetic data can be a valuable choice to increase the amount of available data, be it because of cost or time restrictions or to improve the safety of all people involved.

\subsection{Synthetic Data}

Synthetic data refers to artificially generated data produced by computer programs that simulate real-world data \cite{assefa2020generating}. It is becoming increasingly important to neural networks due to the limitations of real-world data, such as limited quantity, quality, and diversity. Data collection is a significant bottleneck in machine learning, and synthetic data can help overcome this limitation.

Artificial data can be generated at a lower cost and faster rate than real-world data, making it more feasible for large-scale training of neural networks \cite{seib2020mixing}. It can handle the variability in real-world data by randomizing the simulator's parameters, such as lighting, pose, and object textures, to force the neural network to learn the essential features of the object of interest. Augmenting real-world data by generating additional samples completing the data set \cite{wang2019learning} is also possible.

Synthetic data has been used in various applications of neural networks in several fields of knowledge, such as natural scene text recognition \cite{jaderberg2014synthetic}, three-dimensional nuclear segmentation of biological images \cite{dunn2019deepsynth}, and automatic license plate recognition \cite{bjorklund2017automatic}. 

However, the effectiveness of synthetic data depends on how well it represents real-world data and how well it generalizes to new data \cite{goswami2020impact}. Therefore, it is vital to design and evaluate synthetic data generation methods carefully and to combine them with real-world data in a balanced way to achieve optimal performance of neural networks.

\subsection{Simulators}

Simulators are essential in transport engineering, allowing engineers to test and evaluate new designs and technologies in a safe and controlled environment. They replicate real-world scenarios, enabling engineers to study and predict the behavior of transportation systems under different conditions. Simulators can be used to test new vehicle designs \cite{ventura2022high}, to model driver's behavior \cite{de2020perception}, to develop traffic management systems \cite{rizwan2022simulation}, and can help engineers identify potential problems and optimize performance before deploying new systems in the real world.

Video games are increasingly being used as a source of data to train neural networks applied to transport engineering and are viewed as simulators. In \cite{wicaksana2022model}, the authors use a city builder game to visualize urban modeling techniques. In \cite{wurman2022outracing}, a video game is used to train and test AI drivers' capabilities, being able to supplement their human counterparts.

\begin{figure}[h]
\centering
\includegraphics[width=\textwidth]{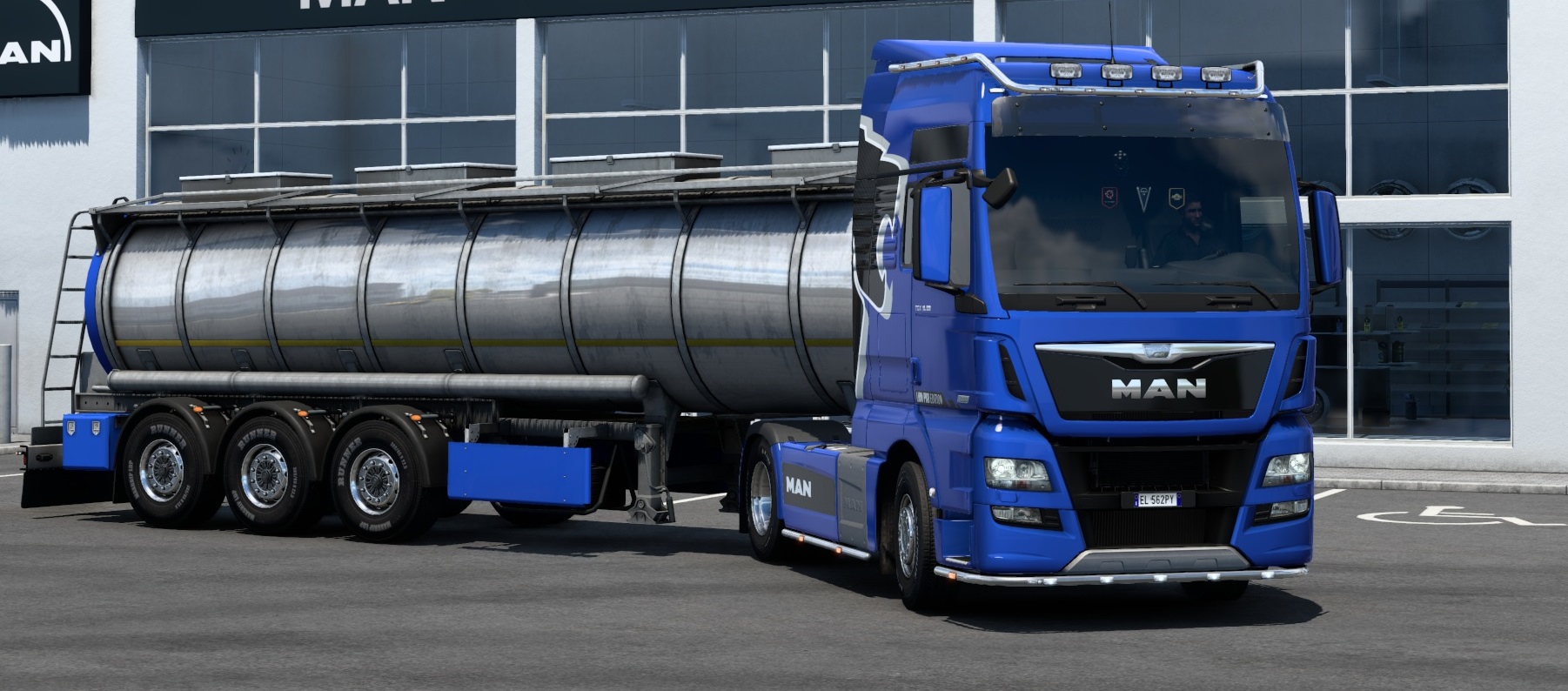}
\caption{3D model of a truck from the video game Euro Truck Simulator 2.}
\label{fig:real}
\end{figure}

Euro Truck Simulator 2 \cite{eurotruck2}, a video game that simulates the experience of driving a truck across Europe, can be a valuable resource for gathering truck images and data for the field of transportation engineering. The game's simulation of different driving conditions and scenarios could provide data on truck performance and driver behavior, informing the design of safer and more efficient transportation systems for heavy vehicles. Additionally, the high level of detail and realism in replicating various truck models, road networks, and driving conditions could provide a rich dataset for analysis. Figure~\ref{fig:real} shows a realistic model of a truck extracted from the game.

To the best of our knowledge, there are no previous works using video games as a source of synthetic images to train a truck axle detection neural network. This gap presents a unique opportunity for research, as leveraging synthetic truck data from video games could provide a cost-effective and scalable solution for generating diverse training datasets, particularly in scenarios where real-world truck images are limited or difficult to obtain.

\section{Experiment}

\begin{figure}[h]
\centering
\includegraphics[width=\textwidth]{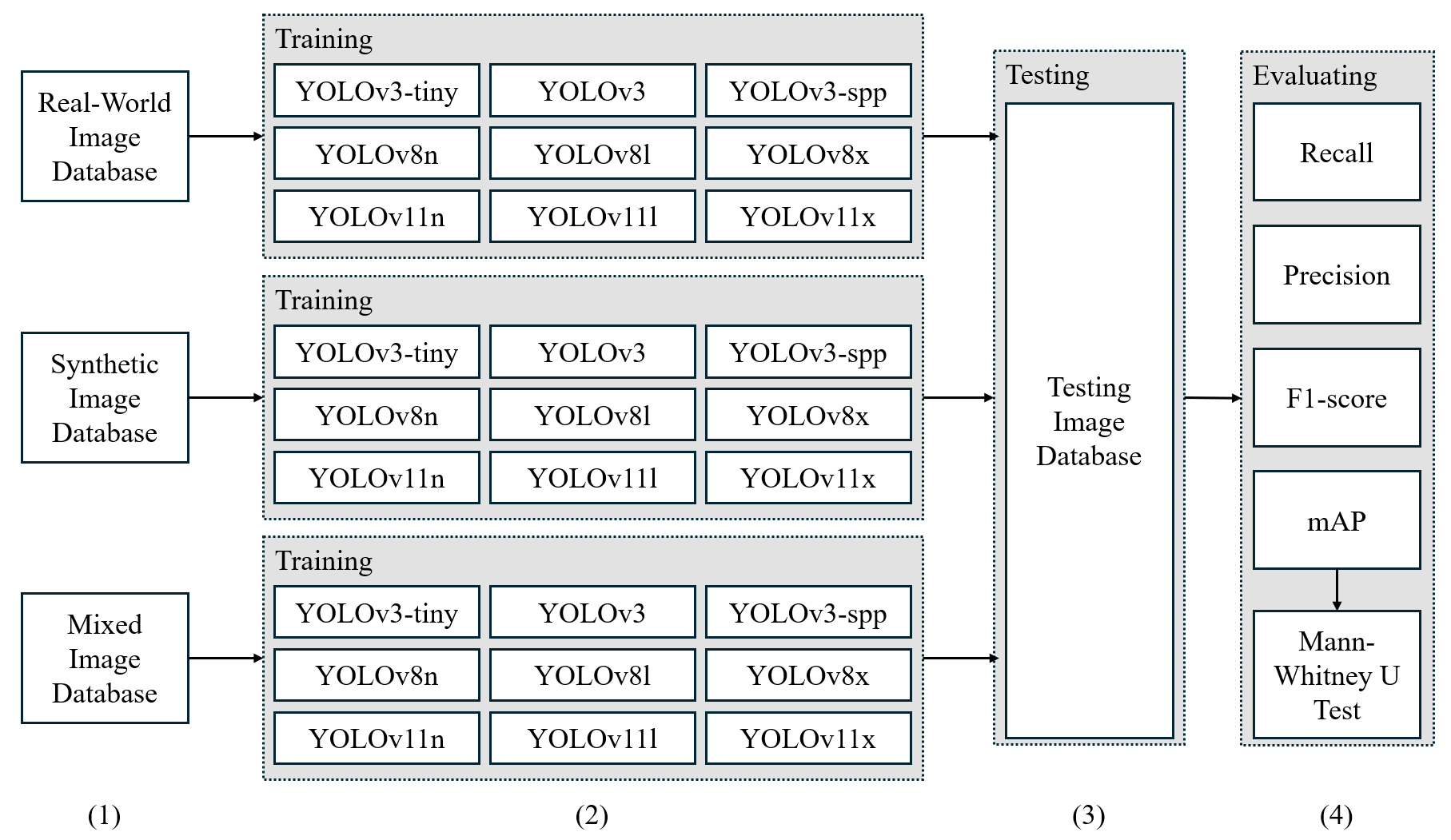}
\caption{All four steps of the experiment, from the creation of the database, (1), to the evaluation of all metrics, (4).}
\label{fig:scheme}
\end{figure}

Since YOLO is a stable algorithm developed for several years, its parameters are well-studied and understood. Because of that, it was chosen as a base to test the ability to use synthetic data extracted from Euro Truck Simulator 2 to train a neural network. In order to compare results and fulfill the objective, the experiment was divided into four parts: (1) building a database of images, (2) training three variations of the YOLO algorithm, each with three different base models, (3) applying the models on a fixed number of images, (4) evaluating the results with five different metrics. Figure~\ref{fig:scheme} shows the method applied to this paper.

\subsection{Database}

For this paper, four different databases were created. It is important to note that all databases used for training purposes must have a similar number of images and objects of interest to eliminate interference from any variables that should not influence the results. However, the database used during the testing phase may contain a different number of images as long as it is kept the same for all models to compare results. Furthermore, the objects of interest in this paper are truck axles. Because of that, truck images where axles are visible were selected.

\begin{figure*}[h]
	\begin{minipage}[t]{\textwidth}		
		\centering
		\includegraphics[width=0.985\textwidth]{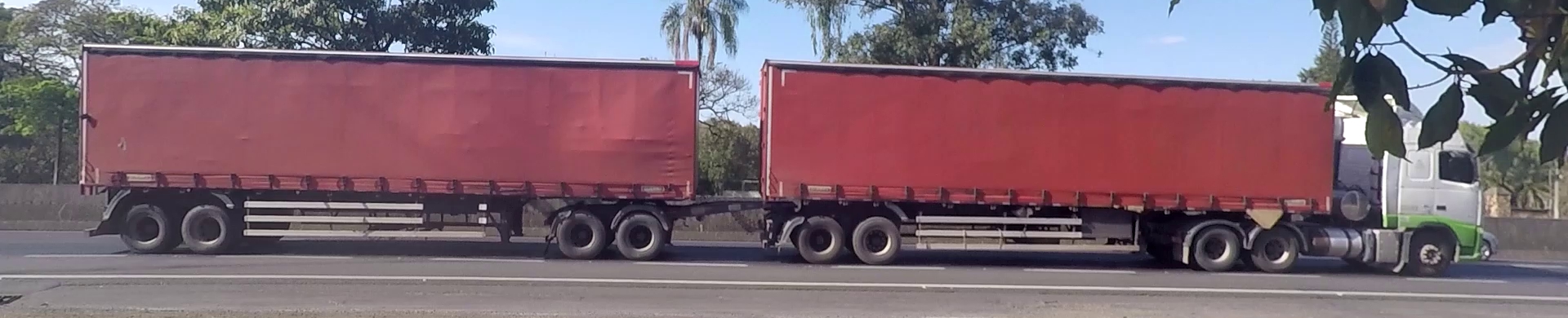}
		\label{fig:sub:subfigure1a}
	\end{minipage}
	\begin{minipage}[t]{\textwidth}
		\centering
		\includegraphics[width=0.985\textwidth]{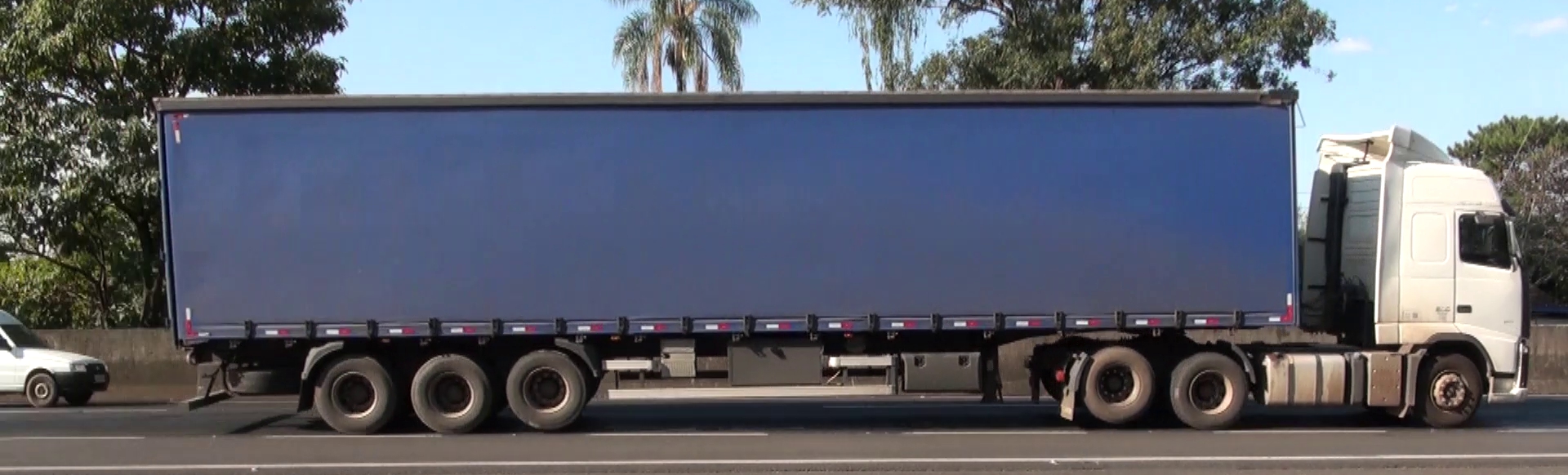}
		\label{fig:sub:subfigure1b}
	\end{minipage}
    \begin{minipage}[t]{\textwidth}
        \centering
        \includegraphics[width=0.985\textwidth]{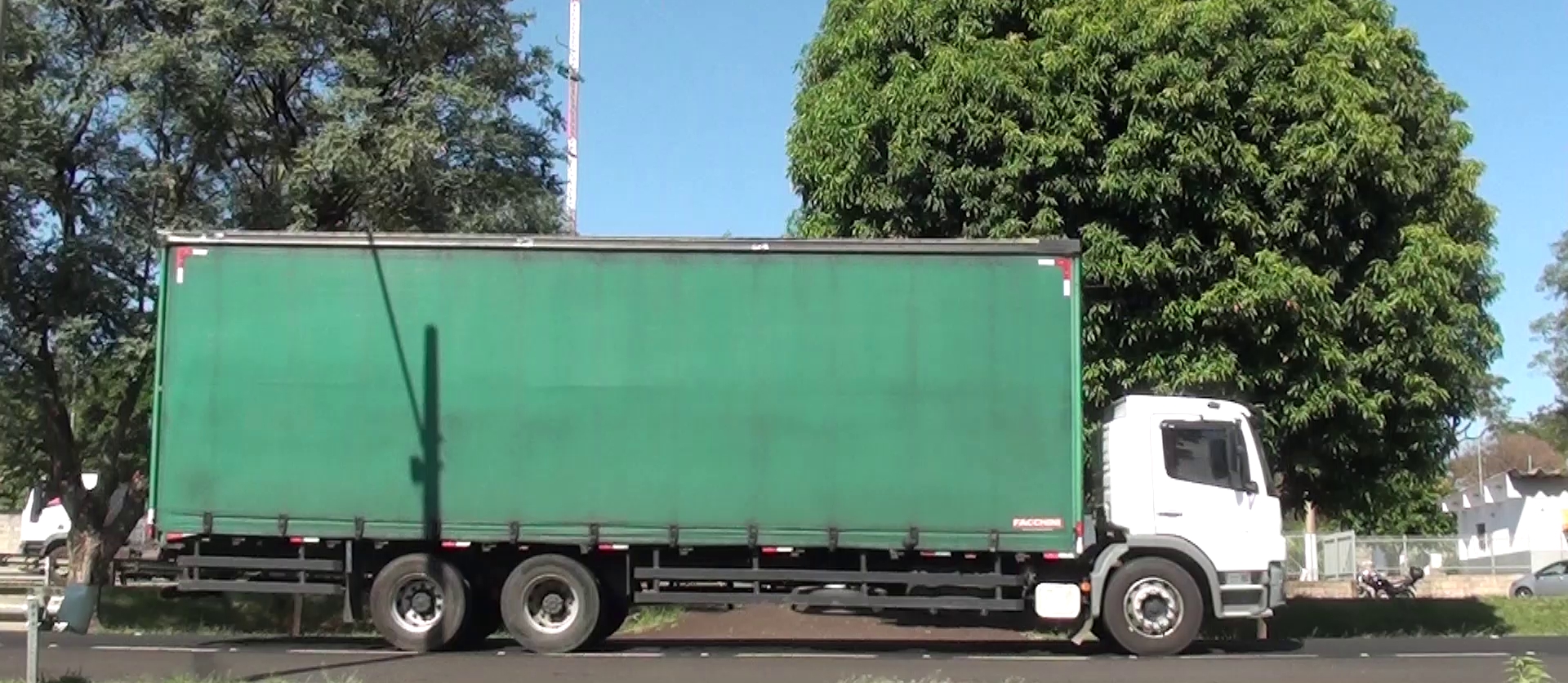}
        \label{fig:sub:subfigure1c}
    \end{minipage}
	\caption{Example of trucks from the first database containing real-life trucks.}
	\label{fig:ex1}
\end{figure*}

The first database contains 346 real-world images of trucks extracted from videos on two different highways in Brazil. The videos were recorded on separate days, including a variation of weather and lighting conditions, with the camera positioned on the side of the road, on the right-side lateral clearance, as seen in Figure~\ref{fig:ex1}.

\begin{figure*}[h]
	\begin{minipage}[t]{\textwidth}		
		\centering
		\includegraphics[width=0.985\textwidth]{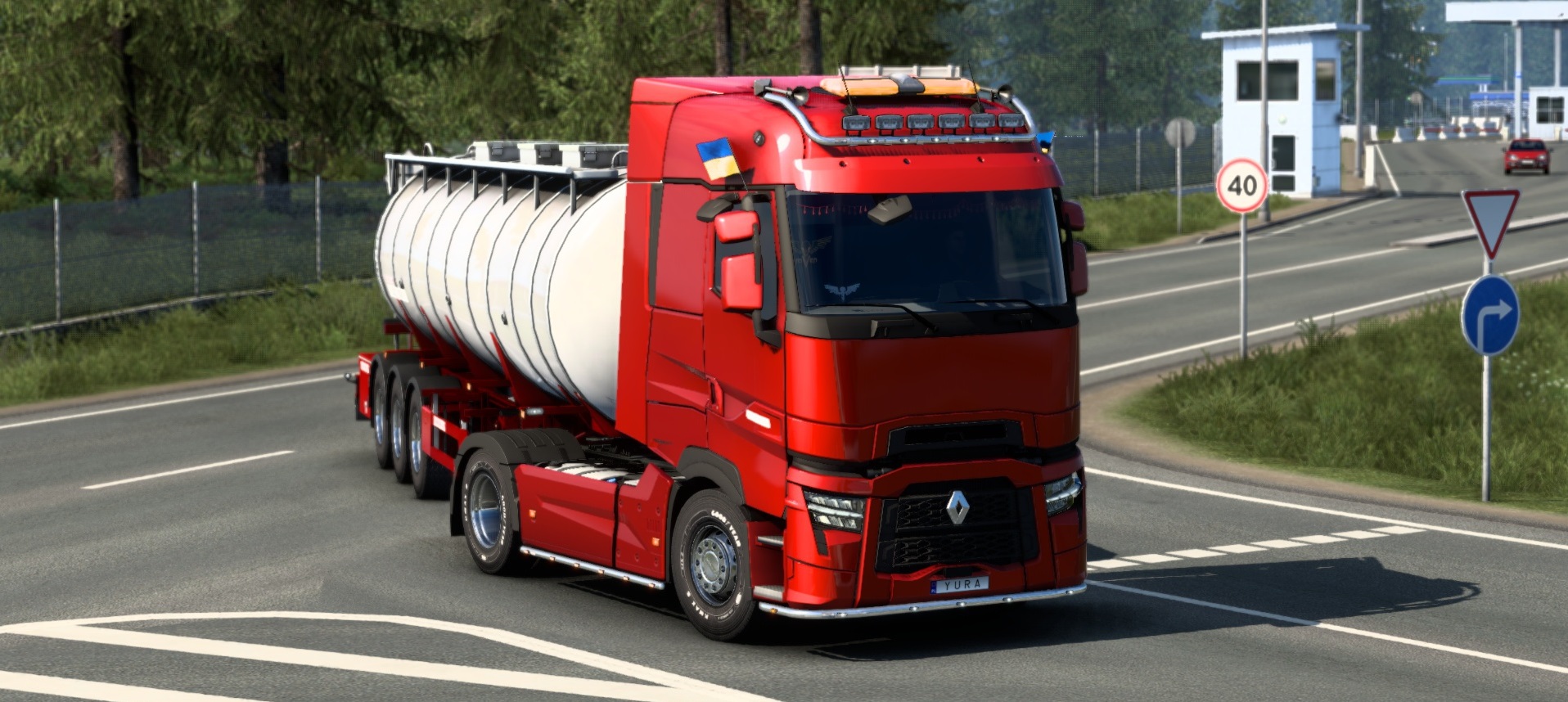}
		\label{fig:sub:subfigure1aa}
	\end{minipage}
	\begin{minipage}[t]{\textwidth}
		\centering
		\includegraphics[width=0.985\textwidth]{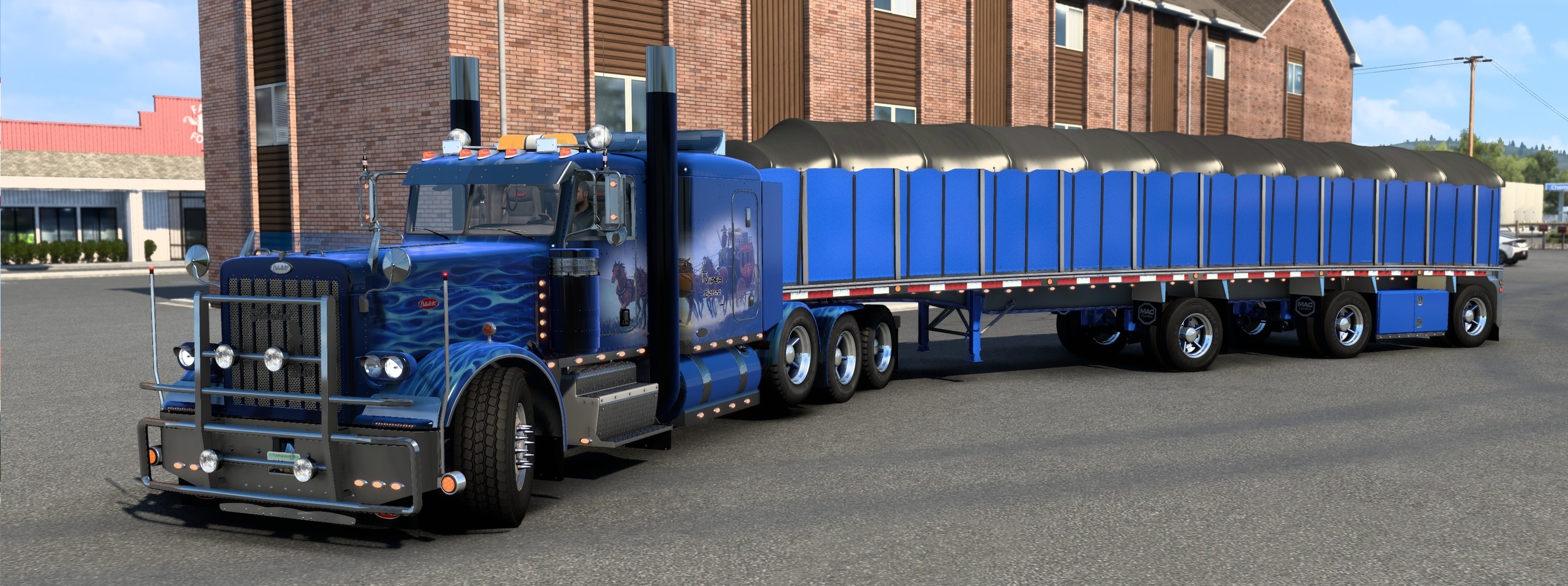}
		\label{fig:sub:subfigure1bb}
	\end{minipage}
	\caption{Example of trucks from the second database containing synthetic images with 3D models.}
	\label{fig:ex2}
\end{figure*}

The second database contains 326 synthetic images from the Euro Truck Simulator 2 video game, which are available on the community site for the game \cite{worldoftrucks}. The images were extracted and selected to contain at least one axle in plain view so the neural network could extract the necessary information. Images are usually artistic since people mostly use the community site to showcase their trucks. Even though that is the case, truck models are depicted in realistic form, and truck axles are mostly visible. Examples of trucks used in the database can be seen in Figure~\ref{fig:ex2}.

The third database was constructed with mixed images extracted from both previous databases, with 175 images from real-life trucks and 175 from synthetic 3D trucks. The number of images was chosen to maintain a similar number of truck axles in all databases, with approximately the same number of images.

The fourth database, the testing set of images, was built using only images from the real world, all different from those in the first database since the application is intended to detect truck axles in real-life truck images. It is composed of 36 images with 119 truck axles. A summary of all databases can be seen in Table~\ref{table:tab1}.

\begin{table}[h]
	\centering
	\caption{Training and testing databases, with the number of images and axles.}
	\label{table:tab1}
	\begin{tabular}{ccccc} 
	    \toprule
		& \begin{tabular}[c]{@{}c@{}} \textbf{First Database}\\\textbf{Real-life Trucks}\end{tabular} & \begin{tabular}[c]{@{}c@{}} \textbf{Second Database}\\\textbf{Synthetic Trucks}\end{tabular} & \begin{tabular}[c]{@{}c@{}} \textbf{Third Database}\\\textbf{Mixed Trucks}\end{tabular} & \textbf{Testing Database}\\ 
		\midrule
		\textbf{Images} & 346  & 326 & 350 & 36\\
		\textbf{Axles}  & 1184 & 1148 & 1176 & 119\\
		\bottomrule
	\end{tabular}
\end{table}

\subsection{Training}

It is common to use pre-trained weight files as a starting point for all neural networks to enhance the efficiency of the training process. For the YOLOv3, YOLOv8 and YOLOv11 architectures used in this paper, three specific pre-trained weight files were utilized for each version: YOLOv3-tiny, YOLOv3, and YOLOv3-spp, YOLOv8n, YOLOv8l and YOLOv8x, and YOLOv11n, YOLOv11l and YOLOv11x. Incorporating these pre-trained weights as a base can expedite the training procedure while benefiting from the knowledge encoded in these models.

YOLOv3-tiny is a lightweight variant optimized for faster inference on resource-constrained devices, with 23 layers and approximately 8.7 million parameters \cite{maya2023pedestrian}. Compared with the standard YOLOv3, it has fewer layers and parameters, resulting in smaller model size and quicker inference speed \cite{YI201917}. However, YOLOv3-tiny sacrifices some accuracy in object detection, particularly for small objects or objects at longer distances.

YOLOv3 is the standard version of the applied YOLO algorithm. It is deeper and more complex than YOLOv3-tiny, with 106 layers and around 61 million parameters \cite{redmon2018yolov3}. YOLOv3 performs well across different object sizes and can accurately detect a wide range of objects. However, due to its larger size, it is slower during inference.

YOLOv3-spp is an enhanced version of YOLOv3, with small changes in the number of parameters (63 million) and layers (110), incorporating Spatial Pyramid Pooling (SPP). This module allows the network to capture information at multiple scales by dividing the input image into grids and pooling features separately from each grid \cite{Huang_2020}. YOLOv3-spp performs better in handling objects of various sizes compared to YOLOv3. The SPP module in YOLOv3-spp enhances spatial reasoning and enables accurate detection of small objects.

The differences between YOLOv8n (Nano), YOLOv8l (Large), and YOLOv8x (Extra Large) lie in their model size, computational complexity, and accuracy. YOLOv8n is the smallest and fastest variant, designed for real-time applications on edge devices with limited processing power. YOLOv8l is a more balanced model, providing significantly better detection performance while still maintaining reasonable inference speeds. YOLOv8x, the largest model, maximizes accuracy, but it comes at the cost of slower processing speeds and higher computational demands \cite{yolov8}.

The same happens for YOLOv11. The differences lies on the size of the network and the number os parameters used, following the same pattern as for YOLOv8 \cite{yolov11}.

To achieve the objective of this paper, the three different databases created for this paper were used to train each of the different base weight files, resulting in 27 separate neural networks, all capable of detecting truck axles in truck images. Then, by a comparative analysis of the results, it was possible to evaluate the feasibility of employing synthetic images as training data for neural networks and assess their impact on the efficacy of truck axle detection.

\subsection{Evaluation}

The choice of performance metrics in truck axle detection depends on the relative costs associated with false positives and false negatives. When the costs attributed to false positive outcomes are substantial, 'precision' serves as a valuable comparison metric. On the other hand, if the costs associated with false negatives are higher, 'recall' emerges as a suitable metric. Misclassifying a truck axle as either a false positive or a false negative could lead to incorrect categorization, resulting in erroneous fees and weight limit determinations, for example. Given its ability to balance precision and recall, the 'F1-Score' assumes significance as an essential metric for evaluation and comparison purposes. Additionally, the 'Mean Average Precision' (mAP) metric is employed to assess the models' object detection capabilities, specifically discerning how they accurately localize the target area. Therefore, mAP was also included in the analytical assessment.

To evaluate the statistical significance of the results from all models, the Mann-Whitney U Test was applied. The Mann-Whitney test is suitable for comparing two independent samples, such as the mAP results of two distinct models, without assuming normality or homogeneity of variances \cite{mann1947test}. Additionally, the Mann-Whitney test is robust to outliers and variations in data scale, which is particularly relevant in computer vision tasks, where results can be influenced by factors such as lighting, occlusion, or dataset variations. This robustness ensures that comparisons remain reliable even in scenarios where data exhibit high variability.

\subsection{Experimental Results}

\begin{figure}[h]
    \centering
	\begin{subfigure}[b]{0.985\textwidth}		
		\centering
		\includegraphics[width=\textwidth]{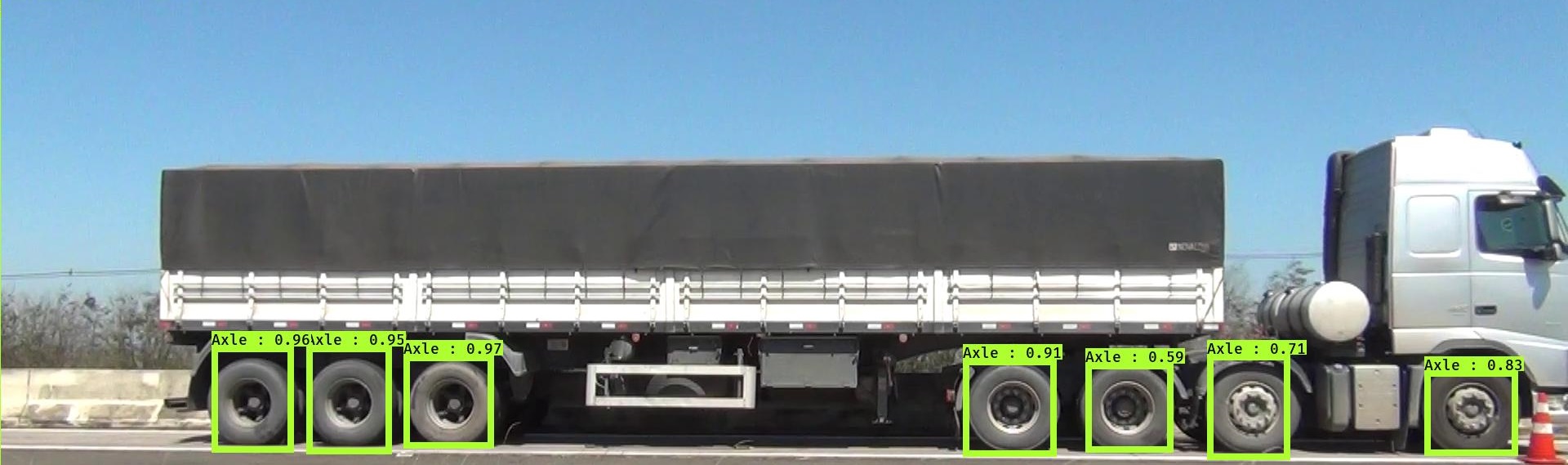}
		\caption{}
		\label{fig:sub:subfiga}
	\end{subfigure}
	\begin{subfigure}[b]{0.985\textwidth}		
		\centering
		\includegraphics[width=\textwidth]{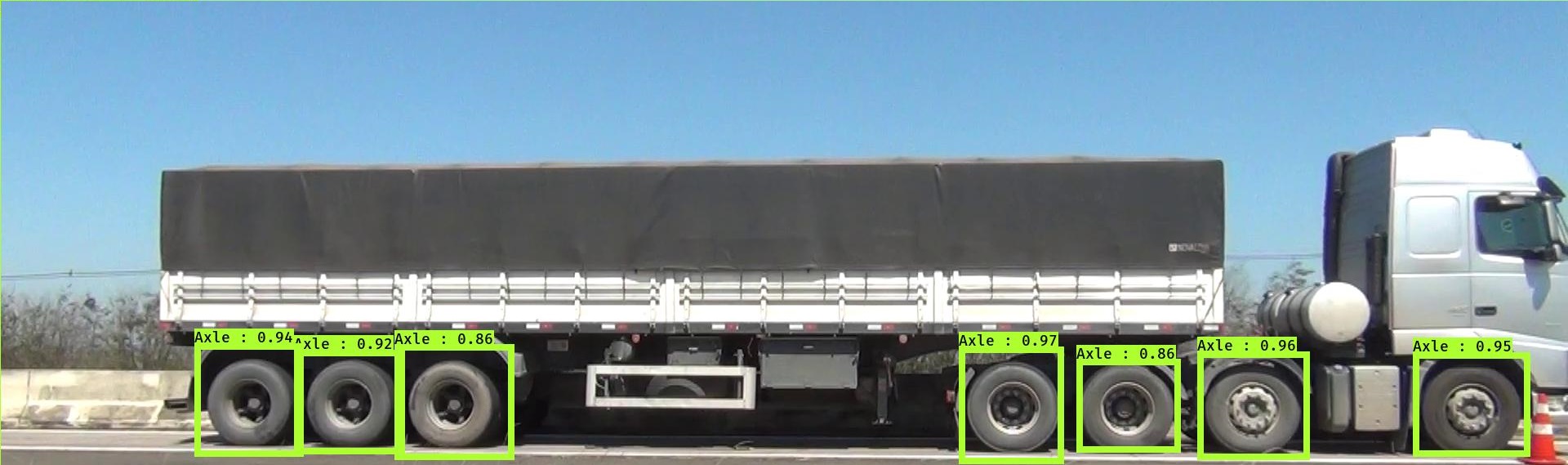}
		\caption{}
		\label{fig:sub:subfigb}
	\end{subfigure}
	\begin{subfigure}[b]{0.985\textwidth}		
		\centering
		\includegraphics[width=\textwidth]{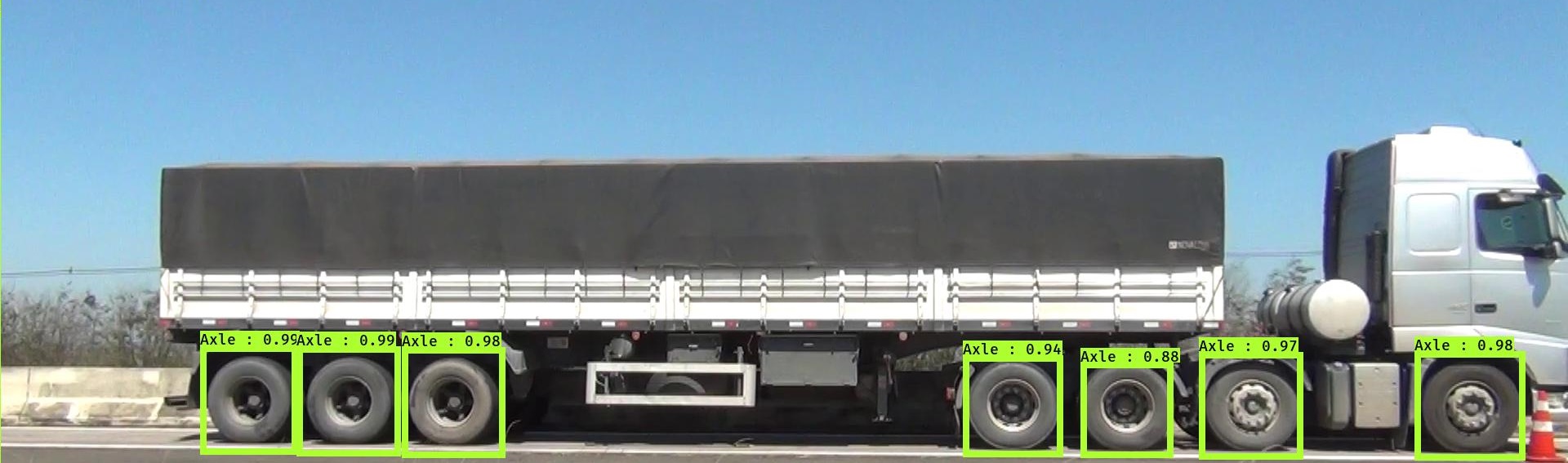}
		\caption{}
	\label{fig:sub:subfigc}
	\end{subfigure}
	\caption{Results from the YOLOv3-spp network, with truck axle detections, trained on the three available databases  (a) Real Trucks (b) Synthetic Trucks (c) Mixed Real and Synthetic Trucks.}
	\label{fig:detectionresults}
\end{figure}

The experiment involved utilizing 36 distinct images, different from the training set, and encompassing a total of 119 truck axles. As all the images were obtained from actual road traffic, they captured various wheel and hubcap configurations. The algorithms were developed to leverage the processing capabilities of the video board, specifically a GeForce GTX 1080. They were implemented using Python programming language, incorporating TensorFlow and Keras packages. Figure \ref{fig:detectionresults} shows an example of output detection images.

The resulting 27 neural network models were evaluated by comparing precision, recall, F1-score, and mAP. These metrics are used to measure the performance related to finding and classifying objects, which is the paper's main focus of comparison. 

\section{Results and Discussions}

The detection results, generated by training and applying all models to the testing dataset, can be seen in Table~\ref{table:tab2}. The results only show the number of axles detected, with false positives and negatives. Based on those numbers, it is possible to notice that YOLOv3-tiny benefited the most from using the mixed image dataset with both synthetic and real trucks. 

The other models presented consistent performance across all tests, indicating that integrating synthetic images effectively expands the diversity and quantity of available data for training purposes. This also suggests the potential of using synthetic images from a video game to enhance machine learning models' overall performance and robustness, providing valuable insights into the advancement of data augmentation techniques and their applicability in various domains.

\begin{table}[h]
\centering
\caption{Detection results for truck axles on all tested neural networks.}
\label{table:tab2}
\begin{tabular}{ccccc}
\toprule
Training Database                                  & Model       & True Positive & False Positive & False Negative \\ \midrule
\multicolumn{1}{c}{\multirow{9}{*}{Real Trucks}} & YOLOv3-tiny & 2             & 153            & 117            \\ 
                                                 & YOLOv3      & 111           & 11             & 8              \\ 
                                                 & YOLOv3-spp  & 117           & 2              & 2              \\ 
                                                 & YOLOv8n     & 88            & 9              & 31             \\
                                                 & YOLOv8l     & 112           & 6              & 7              \\
                                                 & YOLOv8x     & 118           & 2              & 1              \\
                                                 & YOLOv11n    & 91            & 4              & 28             \\
                                                 & YOLOv11l    & 111           & 1              & 8              \\
                                                 & YOLOv11x    & 118           & 1              & 1              \\
\midrule
\multirow{9}{*}{Synthetic Trucks}                & YOLOv3-tiny & 6             & 36             & 113            \\ 
                                                 & YOLOv3      & 118           & 3              & 1              \\ 
                                                 & YOLOv3-spp  & 117           & 1              & 2              \\
                                                 & YOLOv8n     & 92            & 9              & 27             \\
                                                 & YOLOv8l     & 112           & 5              & 7              \\
                                                 & YOLOv8x     & 118           & 1              & 1              \\
                                                 & YOLOv11n    & 92            & 3              & 27             \\
                                                 & YOLOv11l    & 113           & 1              & 6              \\
                                                 & YOLOv11x    & 118           & 1              & 1              \\
\midrule
\multirow{9}{*}{Mixed Trucks}                    & YOLOv3-tiny & 92            & 9              & 27             \\ 
                                                 & YOLOv3      & 116           & 2              & 3              \\
                                                 & YOLOv3-spp  & 117           & 2              & 2              \\
                                                 & YOLOv8n     & 99            & 7              & 20             \\
                                                 & YOLOv8l     & 115           & 2              & 4              \\
                                                 & YOLOv8x     & 119           & 1              & 0              \\
                                                 & YOLOv11n    & 91            & 3              & 28             \\
                                                 & YOLOv11l    & 114           & 1              & 5              \\
                                                 & YOLOv11x    & 119           & 1              & 0              \\
\bottomrule
\end{tabular}
\end{table}

The performance increased when using both images in the mixed database, reaching acceptable detection results for the smaller models. Detections were also better when analyzing the larger models, with increases in true positives and decreases in false positives.

\begin{figure}[h]
\centering
\includegraphics[width=\textwidth]{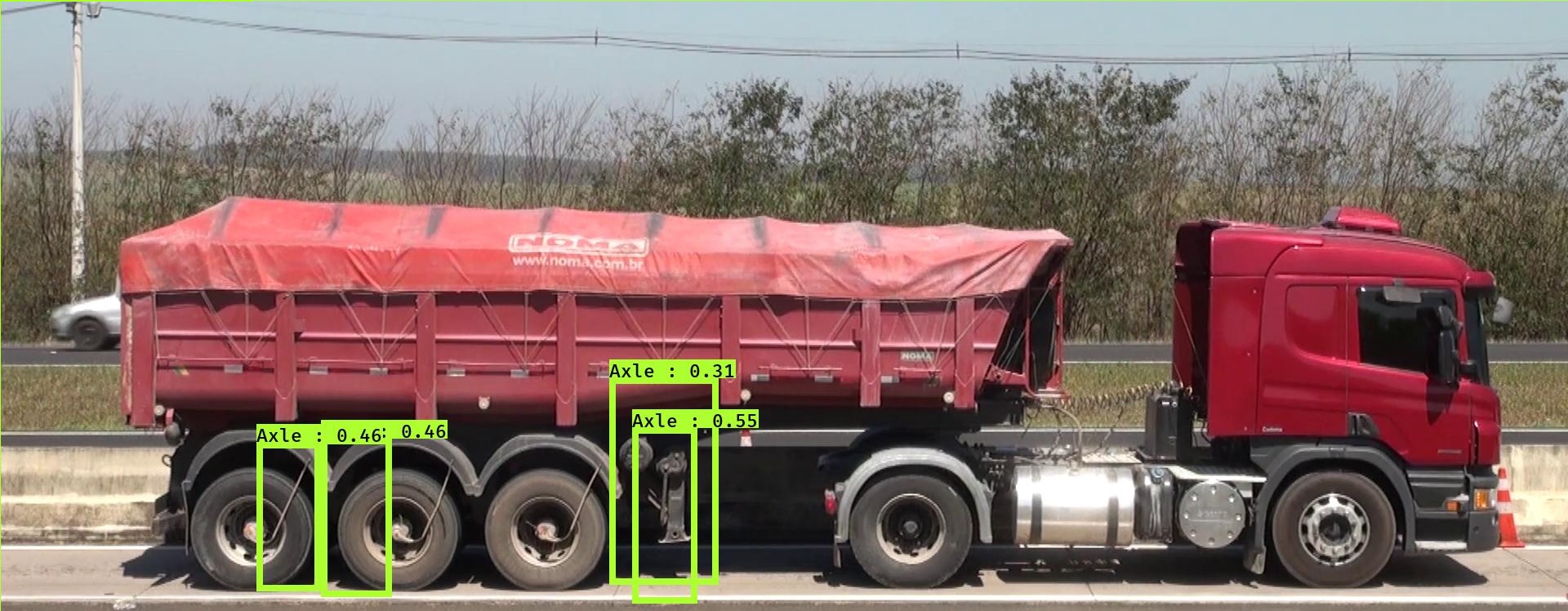}
\caption{False negatives, when truck axles are not detected, and false positives, when objects that are not axles are detected and classified as axles.}
\label{fig:falseposneg}
\end{figure}

The compromise in detection performance for YOLOv3-tiny resulted in many false positives and false negatives when using only real-world or synthetic trucks while training. Looking at the more recent variants of YOLO, the smaller models also had a worse performance when compared with the larger models. It is possible to see examples of false detections in Figure~\ref{fig:falseposneg}.

\begin{figure}
    \begin{subfigure}[b!]{.98\linewidth}
        \includegraphics[width=\linewidth]{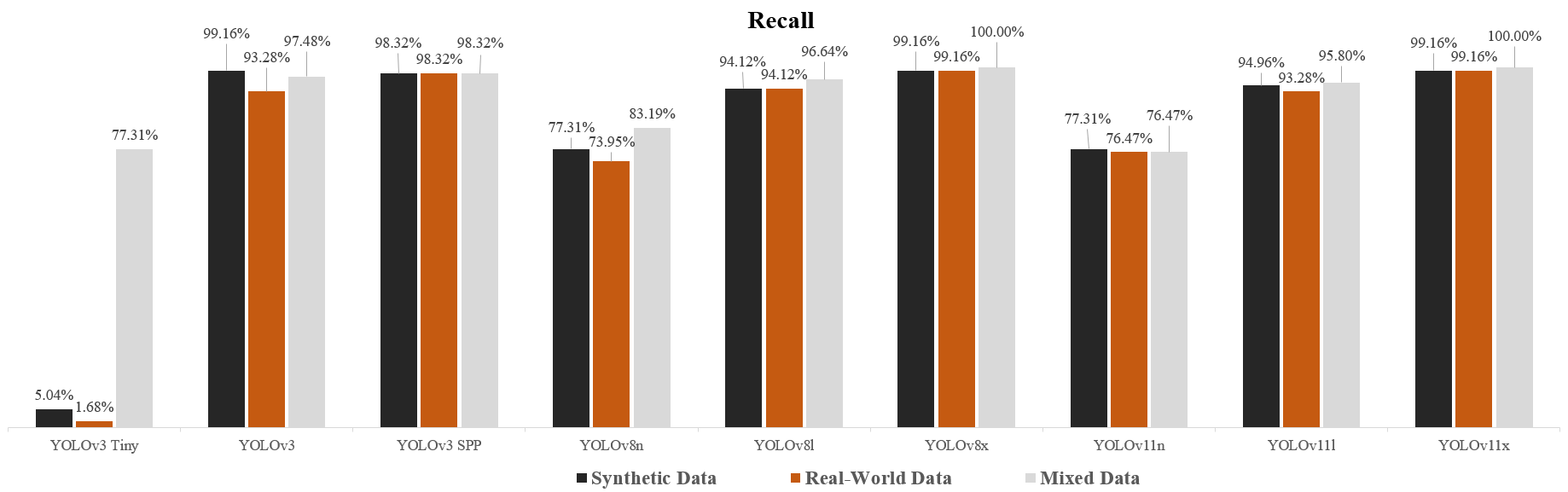}
        \caption{}
    \end{subfigure}
    \begin{subfigure}[b!]{.98\linewidth}
        \includegraphics[width=\linewidth]{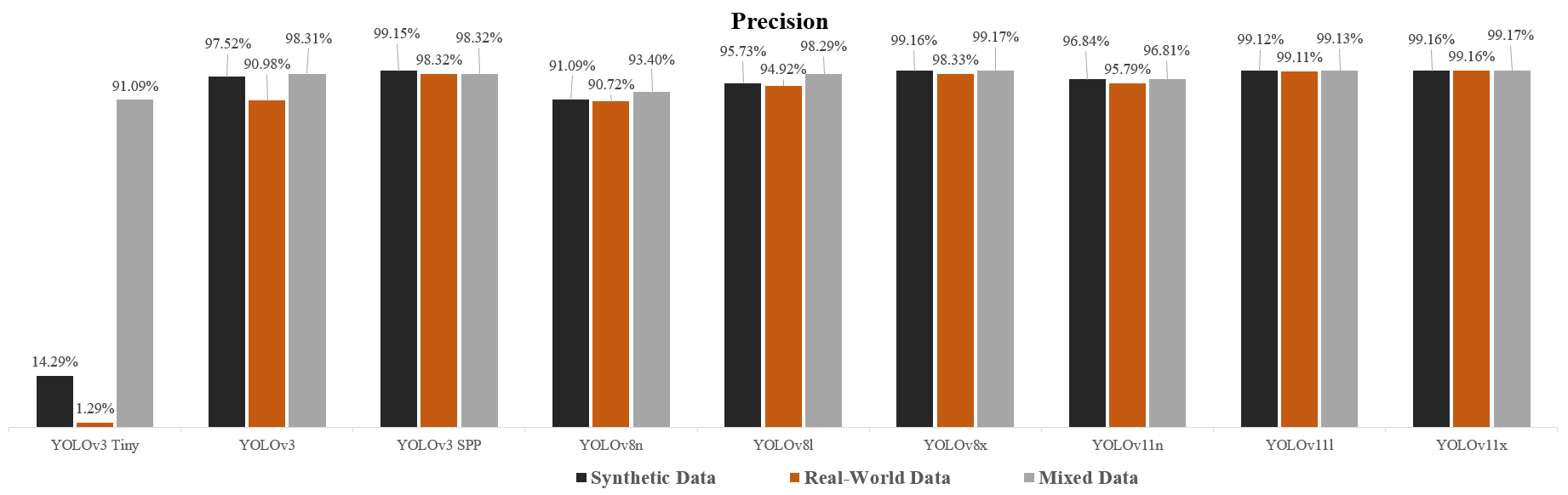}
        \caption{}
    \end{subfigure}
    \begin{subfigure}[b!]{.98\linewidth}
        \includegraphics[width=\linewidth]{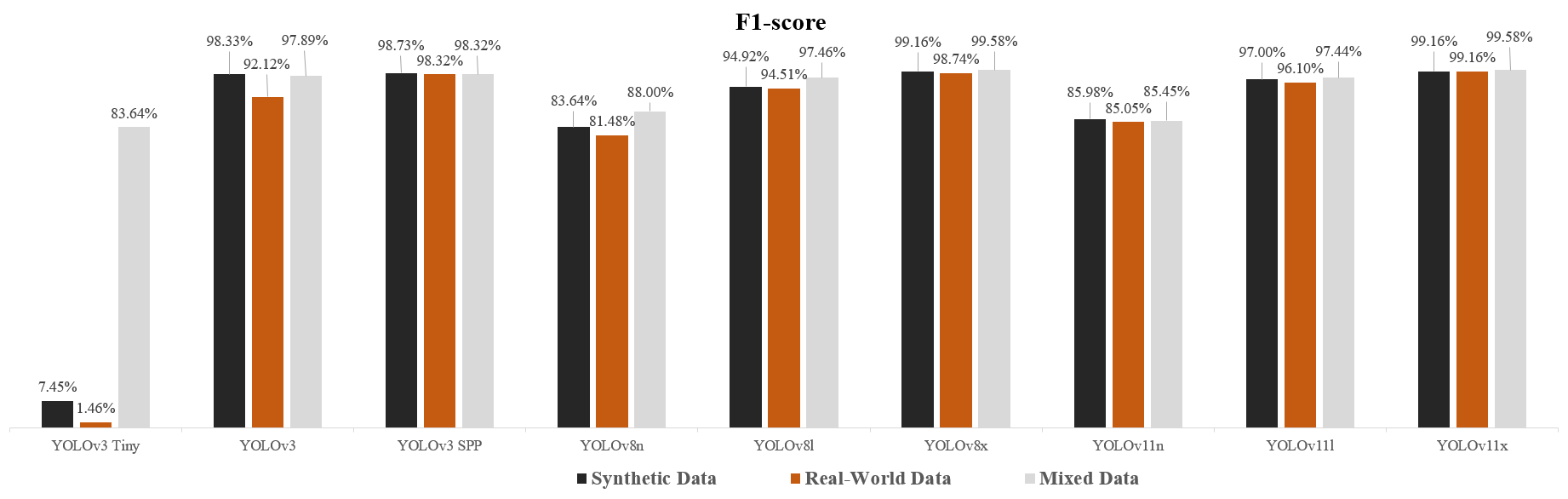}
        \caption{}
    \end{subfigure}
    \begin{subfigure}[b!]{.98\linewidth}
        \includegraphics[width=\linewidth]{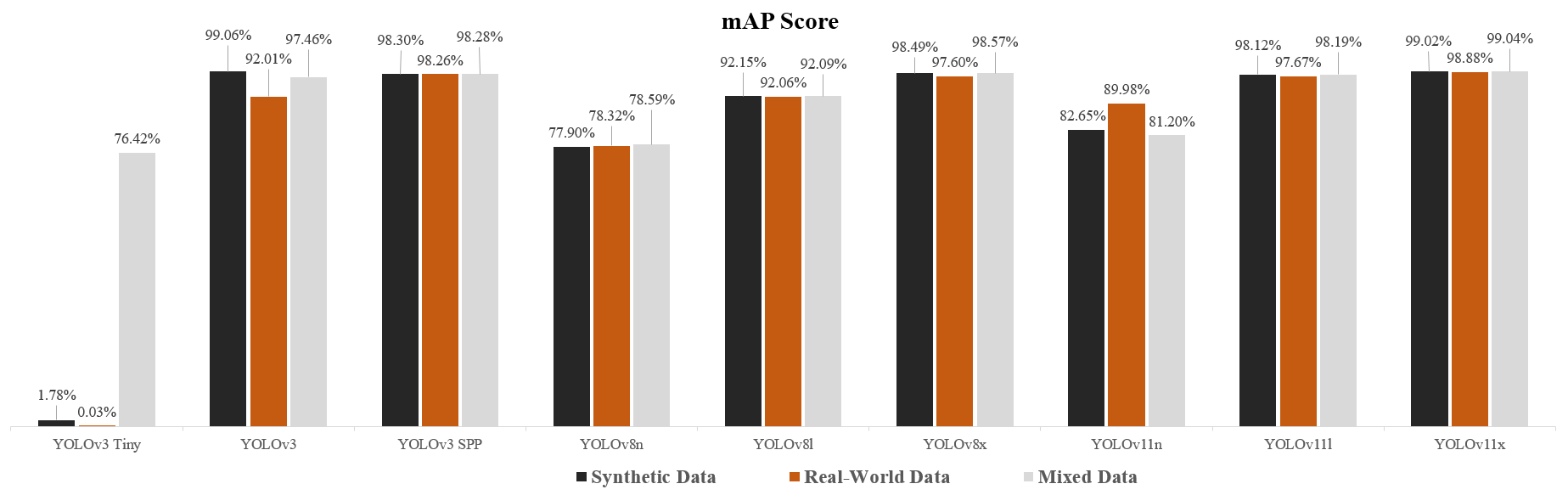}
        \caption{}
    \end{subfigure}
    
    \caption{Graphs comparing all metrics calculated on this paper, in percentages (a) Recall (b) Precision (c) F1-score (d) mAP.}
    \label{fig:graphs}
\end{figure}

With true positives, false positives, and false negatives, it is possible to calculate the metrics used in this paper. The results can be seen in Table~\ref{table:tab3} and visualized in Figure~\ref{fig:graphs}, with graphs. Results are all on percentages, going from 0 to 100\%. The perfect score is 100\%, with all truck axles detected and in the correct position. 

\begin{table}[b!]
\centering
\caption{Calculated metrics for all trained models.}
\label{table:tab3}
\begin{tabular}{cccccc}
\toprule
Training Database                                & Model       & Recall (\%) & Precision (\%)& F1-score (\%)& mAP (\%)\\
\midrule
\multicolumn{1}{c}{\multirow{9}{*}{Real Trucks}} & YOLOv3-tiny & 1.68        & 1.29          & 1.46         & 0.03    \\ 
                                                 & YOLOv3      & 93.28       & 90.98         & 92.12        & 92.01   \\ 
                                                 & YOLOv3-spp  & 98.32       & 98.32         & 98.32        & 98.26   \\
                                                 & YOLOv8n     & 73.95       & 90.72         & 81.48        & 78.32   \\ 
                                                 & YOLOv8l     & 94.12       & 94.92         & 94.51        & 92.06   \\ 
                                                 & YOLOv8x     & 99.16       & 98.33         & 98.74        & 97.60   \\
                                                 & YOLOv11n    & 76.47       & 95.79         & 85.05        & 89.98   \\ 
                                                 & YOLOv11l    & 93.28       & 99.11         & 96.10        & 97.67   \\ 
                                                 & YOLOv11x    & 99.16       & 99.16         & 99.16        & 98.88   \\
\midrule
\multirow{9}{*}{Synthetic Trucks}                & YOLOv3-tiny & 5.04        & 14.29         & 7.45         & 1.78    \\ 
                                                 & YOLOv3      & 99.16       & 97.52         & 98.33        & 99.06   \\ 
                                                 & YOLOv3-spp  & 98.32       & 99.15         & 98.73        & 98.30   \\
                                                 & YOLOv8n     & 77.31       & 91.09         & 83.64        & 77.90   \\ 
                                                 & YOLOv8l     & 94.12       & 95.73         & 94.92        & 92.15   \\ 
                                                 & YOLOv8x     & 99.16       & 99.16         & 99.16        & 98.49   \\
                                                 & YOLOv11n    & 77.31       & 96.84         & 85.98        & 82.65   \\ 
                                                 & YOLOv11l    & 94.96       & 99.12         & 97.00        & 98.12   \\ 
                                                 & YOLOv11x    & 99.16       & 99.16         & 99.16        & 99.02   \\
\midrule
\multirow{9}{*}{Mixed Trucks}                    & YOLOv3-tiny & 77.31       & 91.09         & 83.64        & 76.42   \\ 
                                                 & YOLOv3      & 97.48       & 98.31         & 97.89        & 97.46   \\ 
                                                 & YOLOv3-spp  & 98.32       & 98.32         & 98.32        & 98.28   \\
                                                 & YOLOv8n     & 83.19       & 93.40         & 88.00        & 78.59   \\ 
                                                 & YOLOv8l     & 96.64       & 98.29         & 97.46        & 92.09   \\ 
                                                 & YOLOv8x     & 100.00      & 99.17         & 99.58        & 98.57   \\
                                                 & YOLOv11n    & 76.47       & 96.81         & 85.54        & 81.20   \\ 
                                                 & YOLOv11l    & 95.80       & 99.13         & 97.44        & 98.19   \\ 
                                                 & YOLOv11x    & 100.00      & 99.17         & 99.58        & 99.04   \\
\bottomrule
\end{tabular}
\end{table}

From the models trained with synthetic truck images, YOLOv3-tiny performed poorly with very low mAP (1.78\%) and recall (5.04\%), indicating many missed objects. Additionally, its precision and F1-score were relatively low. In contrast, the YOLOv3 and YOLOv3-spp models demonstrated excellent performance. They achieved high mAP values (99.06\% and 98.30\%, respectively) and recall values (99.16\% and 98.32\%, respectively). These models also had high precision and F1-scores. Like the models trained on synthetic images, YOLOv3 and YOLOv3-spp trained on mixed images performed well. They had high mAP values (97.46\% and 98.28\%, respectively) and recall values (97.48\% and 98.32\%, respectively). These models also demonstrated high precision and F1-scores.

On the other hand, the YOLOv3-tiny model trained on real-world trucks had the worst performance among all the models. It had extremely low mAP (0.03\%), recall (1.68\%), and precision (1.29\%) values. This model missed a significant number of objects and had a high number of false positives. The YOLOv3 and YOLOv3-spp models trained on real-world trucks had relatively high mAP values (92.01\% and 98.26\%, respectively) and good recall values (93.28\% and 98.32\%, respectively). Although they performed well, their performance did not match the models trained on synthetic and mixed truck images.

YOLOv8 followed a similar pattern, with YOLOv8n having a worse performance compared to YOLOv8l and YOLOv8x. It is important to notice that although results were indeed worse, they were still better when compared with YOLOv3tiny. This indicates that the evolution of the smaller models, with reduced number of layers and parameters, are still having an influence on the results, but by a diminishing factor. That's an interesting result, given that smaller models are capable of running on portable devices, contributing to the process of 'IoTzation'.

The same can be said about YOLOv11. All models followed the same pattern of YOLOv8, with an improve in detection results, reaching 100\% recall on the mixed database. Looking at the other metrics, all results on the synthetic and mixed database were above 99\%, indicating an improvement on performance, and giving a hint about the overall result of this paper.

\begin{figure}[H]
    \centering
    \begin{subfigure}[b]{.45\linewidth}
        \includegraphics[width=\linewidth]{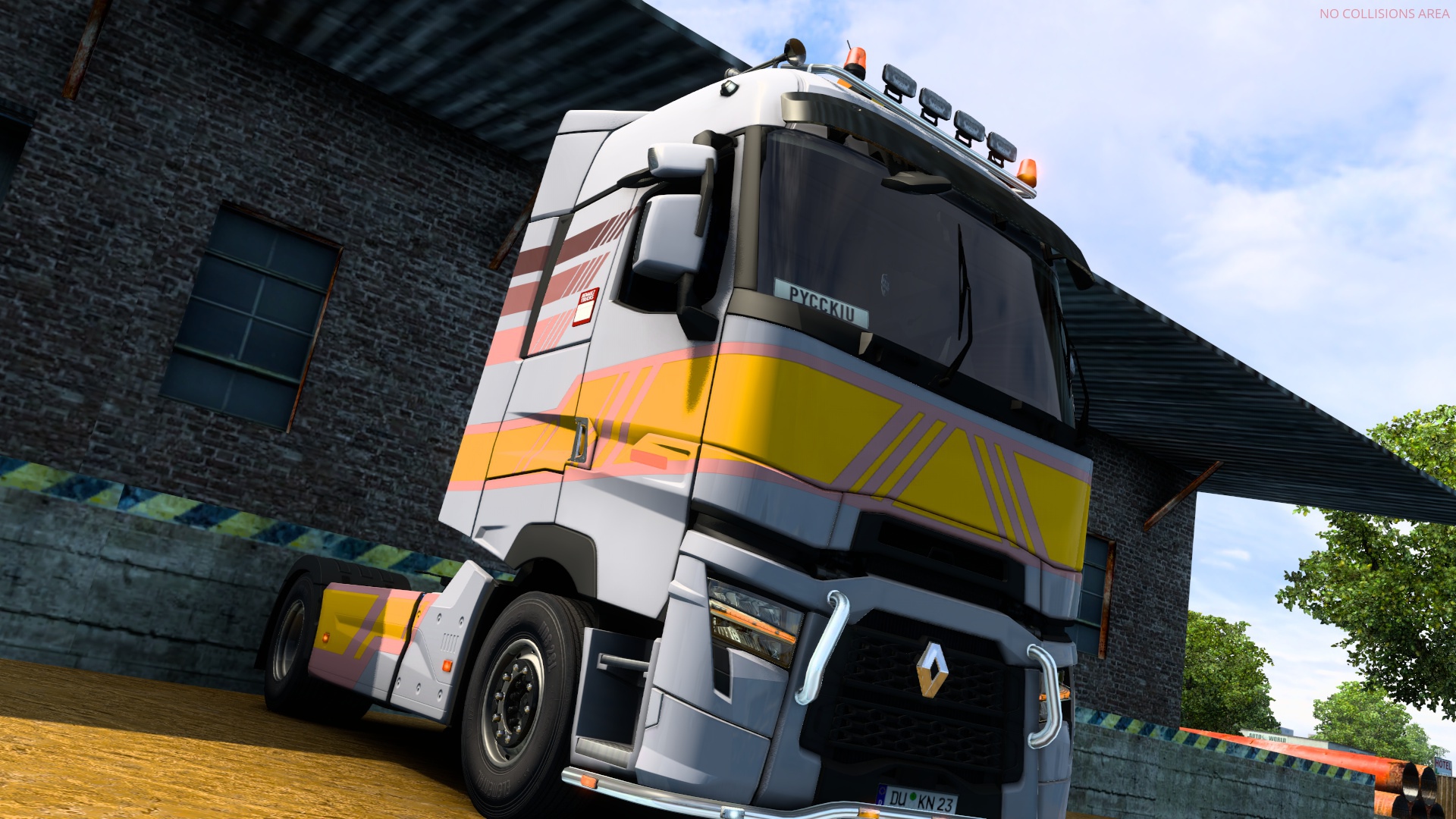}
        \caption{}
    \end{subfigure}
    
    \begin{subfigure}[b]{.45\linewidth}
        \includegraphics[width=\linewidth]{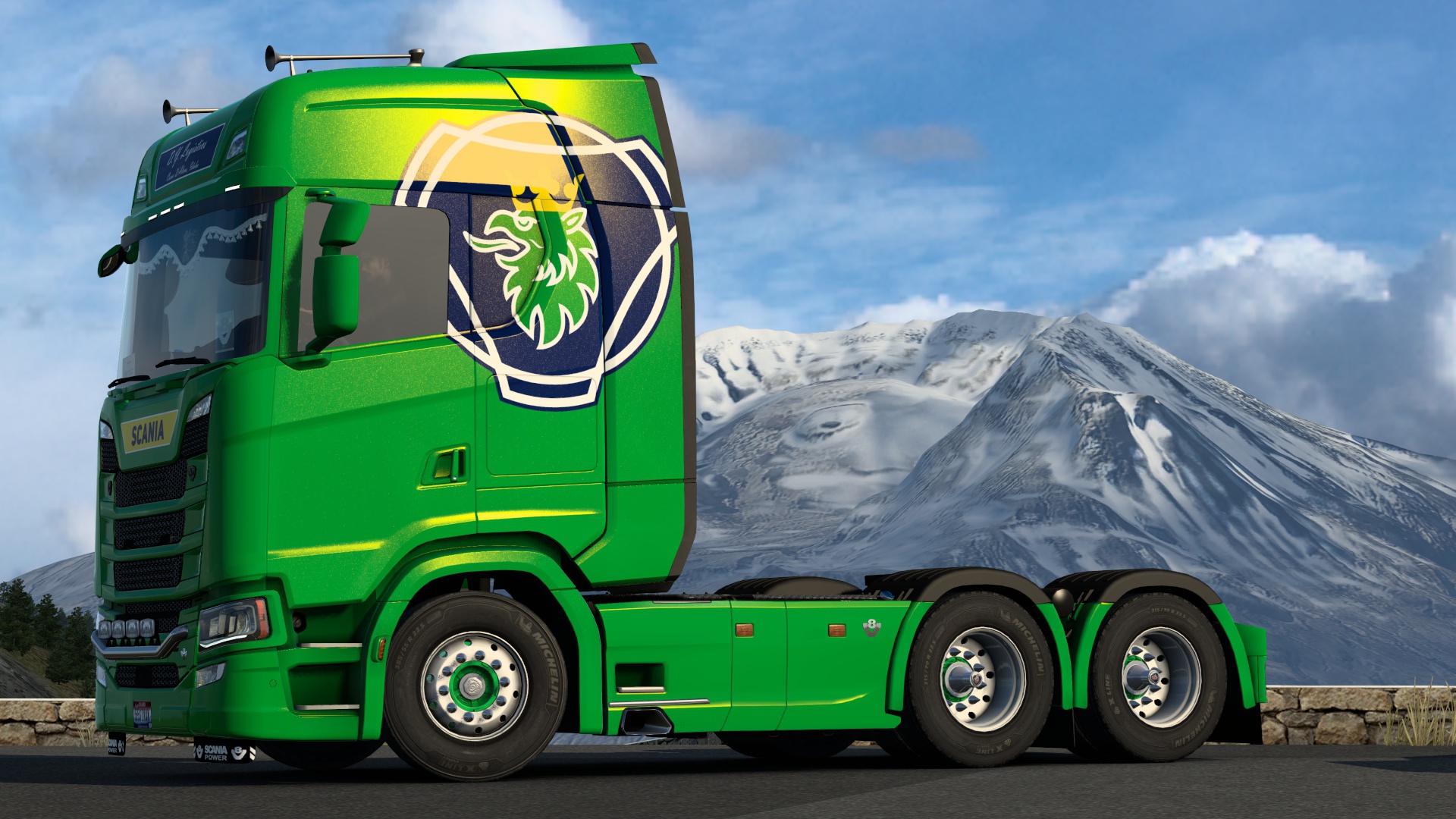}
        \caption{}
    \end{subfigure}
    \begin{subfigure}[b]{.45\linewidth}
        \includegraphics[width=\linewidth]{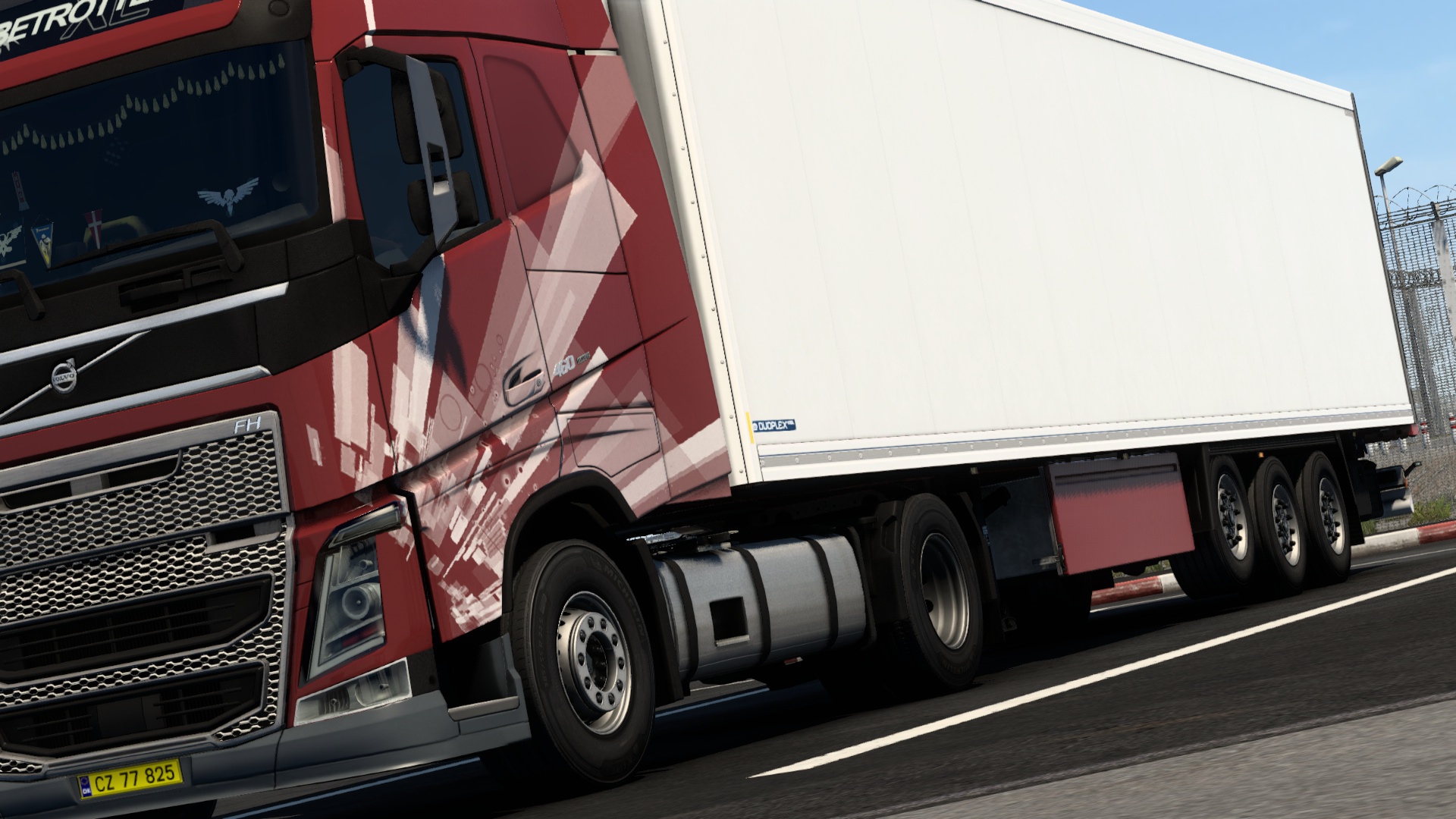}
        \caption{}
    \end{subfigure}
    
    \caption{Variable positions of truck axles in the synthetic image database, providing more information for the neural networks to extract patterns from during the training phase.}
    \label{fig:varpos}
\end{figure}

For all trained models, this paper demonstrated that using a mixed database with synthetic and real-world images meant better or at least comparable results for all trained models. This result might be due to the variability of examples to extract information from during the training phase, which the real-world database lacks. All trucks on the real-world database were from a side viewpoint, perpendicular to the road. Synthetic images have a more comprehensive range of axle positions, as shown in Figure~\ref{fig:varpos}.

The Mann-Whitney U test was applied to the resulting mAP values to determine whether the differences between them are statistically significant or not. Two null hypothesis were constructed and tested.

The first hypothesis, '$H_0:$ There are no differences between neural networks trained on different databases of images.' was created to check if there are any difference between mAp results of neural networks trained on different image databases.

Since there are three possible comparisons to be made, Real Data x Synthetic Data, Real Data x Mixed Data, and Synthetic Data x Mixed Data, the test was applied to each one, with a resulting U value. The U value was then checked against the critical value table, and results are shown on Table~\ref{table:utestdata}.

\begin{table}[b]
\centering
\begin{threeparttable}
\caption{Results from the Mann-Whitney U test for the three image databases.}
\label{table:utestdata}

\begin{tabular}{cccc}
\toprule
                            & U Value & Critical Value* & Results        \\
\midrule
Real Data x Synthetic Data  & 31      & 17             & Fail to Reject \\
Real Data x Mixed Data      & 34      & 17             & Fail to Reject \\
Synthetic Data x Mixed Data & 44      & 17             & Fail to Reject \\
\bottomrule
\end{tabular}
    \begin{tablenotes}[flushleft]
      \small
      \item *$\alpha=0.05, n1=9, n2=9$
    \end{tablenotes}
  \end{threeparttable}
\end{table}

Since the critical value for all tests is 17, we fail to reject the null hypothesis. Therefore, the mAp results distribution from all models are equivalent between each other. This indicates that it is possible to train neural networks using synthetic images without losing performance.

\begin{figure}[h]
    \centering
    \includegraphics[width=0.70\textwidth]{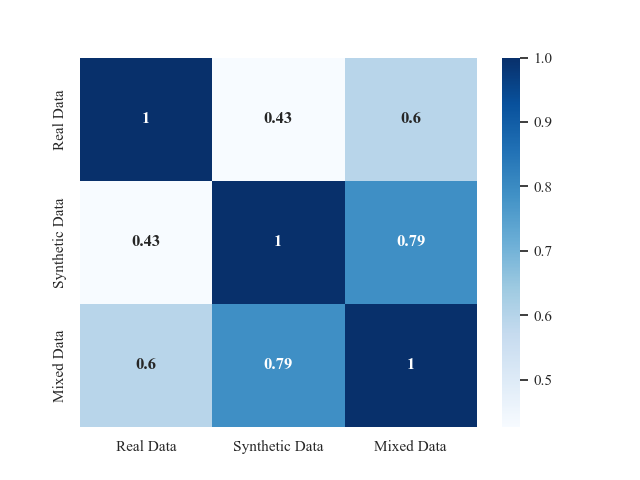}
    \caption{p-values from the Mann-Whitney U test between mAP results from different databases.}
    \label{fig:pvaluedataset}
\end{figure}

It is possible to see the p-values of the Mann-Whitney U tests on Figure~\ref{fig:pvaluedataset}. Results are significant, with p-values above 0.43.

The second hypothesis tested, '$H_0:$ There are no differences on mAP results between versions of YOLO neural networks when trained on different sets of images.' was created to check whether results from the different versions of the YOLO architecture were significant or not on the databases created for this paper. 

Three comparisons are also possible between the versions of the YOLO neural network used on this paper, YOLOv3 x YOLOv8, YOLOv3 x YOLOv11, and YOLOv8 x YOLOv11. All comparisons were also tested with the U test, and results are seen on Table~\ref{table:utestversion}.

\begin{table}[b]
\centering
\begin{threeparttable}
\caption{Results from the Mann-Whitney U test for the three versions of YOLO.}
\label{table:utestversion}

\begin{tabular}{cccc}
\toprule
                  & U Value & Critical Value* & Results        \\
\midrule
YOLOv3 x YOLOv8   & 39      & 17              & Fail to Reject \\
YOLOv3 x YOLOv11  & 33      & 17              & Fail to Reject \\
YOLOv8 x YOLOv11  & 24      & 17              & Fail to Reject \\
\bottomrule
\end{tabular}
    \begin{tablenotes}[flushleft]
      \small
      \item *$\alpha=0.05, n1=9, n2=9$
    \end{tablenotes}
  \end{threeparttable}
\end{table}

With the same resulting critical value of 17, it is not possible to reject the null hypothesis, indicating that the mAP performance between model versions is equivalent. This happens due to the high detection levels of all trained models, since mAp scores were usually above 90\%.

\begin{figure}[h]
    \centering
    \includegraphics[width=0.70\textwidth]{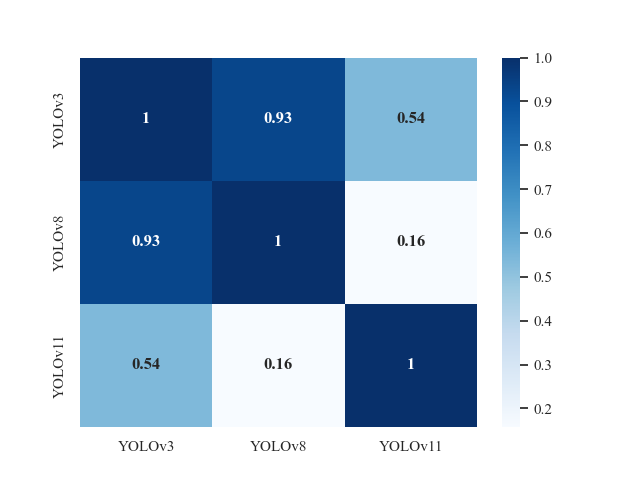}
    \caption{p-values from the Mann-Whitney U test between mAP results from different YOLO versions.}
    \label{fig:pvaluemodels}
\end{figure}

The p-values can be seen on Figure~\ref{fig:pvaluemodels}, with the lowest p-value of 0.16 on the comparison between YOLOv8 and YOLOv11.

Based on all results, it is possible to conclude that using synthetic images from a video game to train neural networks is a viable option, contributing with different examples for extracting knowledge and enriching the dataset. The trained models either improved detection performance or got similar results to training with real-world images. 

Furthermore, extracting the images from the video game used less time, financial, and human resources, with no need to go to the roadside to record trucks. By opting for synthetic images, the risks associated with standing beside a highway, vulnerable to vehicular accidents, exhaust fumes or adverse weather, are effectively negated. Video game images offer the advantage of customization, allowing scenarios to be crafted, increasing variety in examples for the neural network. Thus, prioritizing the integration of video game imagery not only safeguards researchers from potential harm but also ensures the creation of content in a controlled environment.

\section{Conclusion and Future Works}

We tested three different databases, with real-world truck images, synthetic truck images, and mixed images, to train various YOLO networks. The training process took advantage of already learned knowledge of nine base weight files for three distinct sized networks, from smaller to larger: (1) YOLOv3-tiny, YOLOv8n and YOLOv11n, (2) YOLOv3, YOLOv8l, YOLOv11l and (3) YOLOv3-spp, YOLOv8x, YOLOv11x. We evaluated the 27 resulting models using recall, precision, f1-score, and mAP to assess if training a neural network using synthetic images from a video game is possible. Results show that it is possible, with all models performing better or equivalent to neural networks trained with real-world images. This finding is the main contribution of this paper since it opens the possibility to use synthetic images from a video game to enhance the number of examples to train a neural network, saving time and resources.

Mixing images resulted in a better amount of variability, with truck axles in several different angles and lighting conditions. Since neural networks extract knowledge from all examples provided, synthetic images are an excellent choice to enrich a database of images, especially if they come from a video game with realistic graphics. The closer the representation of the object in the synthetic image to real life, the better the results will be.

The results of this paper also showed that even with smaller databases for training and testing, detection and classification were successful. This suggests that using a specialized neural network, specifically trained for a particular task, is a viable alternative to using pre-trained, general-purpose neural networks found online. Moreover, not only is it a good alternative, but it can also be trained with small databases. This is a significant finding because acquiring sufficient data can be challenging for researchers. Therefore, this paper highlights the possibility of using small databases to train specialized neural networks with good results.

To ensure a fair comparison of results, training and testing all networks using the same databases was crucial. Although the training times for each model were significantly longer than the testing times, that was not considered since training is performed only once for each network. Each model required approximately four hours of training, while testing was completed in seconds. While better hardware can reduce training times, larger databases require more processing power for faster processing times. Given that real-world images may be mixed with easily acquired synthetic images, training databases will get larger in future works.

\section{Acknowledgments}

This study was financed in part by the Coordenação de Aperfeiçoamento de Pessoal de Nível Superior - Brasil (CAPES) - Finance Code 001. This research also received financial support from the National Council for Scientific and Technological Development (CNPq), projects 436954/2018-4 and 311964/2022-2.

\bibliographystyle{ieeetr}
\bibliography{ref}

\end{document}